\documentclass{article}

\usepackage[dblblindworkshop,final]{neurips_2026}

\workshoptitle{AXIOM Foundations of Efficient Deep Learning}

\usepackage[utf8]{inputenc}
\usepackage[T1]{fontenc}
\usepackage{hyperref}
\usepackage{url}
\usepackage{booktabs}
\usepackage{amsfonts}
\usepackage{amsmath}
\usepackage{nicefrac}
\usepackage{microtype}
\usepackage{xcolor}
\usepackage{graphicx}
\usepackage{float}
\usepackage{array}

\hypersetup{hidelinks}

\title{What Makes Recurrence Generalize Across Lengths?}
\author{Yoonsung Jung\textsuperscript{1,*}\hspace{0.7em}Sunghun Jung\textsuperscript{1,*}\hspace{0.7em}Louise Nayeong Koh\textsuperscript{1,*}\hspace{0.7em}Yongjun Choi\textsuperscript{1}\\[0.3em]\centering \textsuperscript{1}Yonsei University}

\begin{document}

\maketitle\begingroup\renewcommand{\thefootnote}{*}\footnotetext{Equal contribution.}\endgroup

\begin{abstract}
Multi-step reasoning requires models to preserve and update intermediate information as computation unfolds. State tracking captures this fundamental requirement, but models that succeed within a fixed range of state transitions may not generalize when substantially more updates are required. Recurrent Transformers offer a natural mechanism for addressing this challenge by repeatedly applying shared Transformer blocks, allowing computation to scale with problem length. We study when such recurrent computation generalizes beyond the training range in a controlled state-tracking task, varying operation representation, positional encoding, and recurrent capacity. We find that recurrence alone is insufficient: multi-token template representations and absolute positional encoding can severely limit extrapolation, whereas atomic operation representations enable substantially stronger length generalization. The effect of positional structure also varies with recurrent depth, with periodic positional encoding achieving the strongest mean performance at greater recurrent depth. Most importantly, scaling the recurrent operator extends the extrapolation horizon, reaching 35.1\% exact-match accuracy on sequences of 20–40 operations after training only on 2–10, while substantially larger fixed-depth Transformers remain at 0\%. Additional recurrent computation at test time can extend the range over which the model generalizes, but does not eliminate failure on much longer sequences. These results suggest that recurrent computation can support length-generalizable state tracking, but that its effectiveness depends critically on whether the learned computation remains reusable across sequence lengths.

\end{abstract}

\section{Introduction}

State tracking is a fundamental capability underlying multi-step reasoning, requiring models to maintain and update relevant state across successive operations \citep{dai2024representational,li2025state,tang2026entity}. However, standard Transformers have fixed computational depth as sequence length grows, which may limit tasks requiring increasingly many state updates \citep{merrill2023parallelism,liu2023shortcuts}. Recurrent computation offers a natural alternative by repeatedly applying shared Transformer blocks, allowing computational depth to grow without proportionally increasing parameter count \citep{dehghani2019universal}. Prior work has shown that such recurrent-depth Transformers can improve length or reasoning-depth extrapolation \citep{fan2025looped,kohli2026loop}.

Yet recurrence alone does not ensure that learned computation remains reusable beyond the training range. In particular, prior work has shown that positional encoding can strongly affect Transformer length generalization \citep{kazemnejad2023impact,ruoss2023randomized}. We therefore ask under what representational and positional conditions recurrent computation generalizes across sequence lengths. Using a controlled state-tracking task based on sequential pairwise swaps, we systematically vary computational architecture, operation representation, and positional encoding.

We find that recurrence alone is insufficient: length generalization depends strongly on how operations and positional information are represented. Atomic operation representations substantially improve extrapolation, while absolute positional structure can sharply limit it. Moreover, scaling recurrent capacity extends the extrapolation horizon, whereas scaling fixed-depth Transformers yields substantially weaker OOD gains. These results motivate periodic positional encoding as a simple length-reusable positional bias and suggest that successful recurrent length generalization depends not merely on additional computation, but on whether that computation can be reused across lengths.


\section{Methods}

\subsection{Task, Data, and Representations}
We study length generalization using a synthetic state-tracking task with five entities and five uniquely assigned states, represented as colored balls. Each example begins with a random one-to-one assignment of colors to entities, followed by a sequence of pairwise swaps. The model predicts the complete assignment after all swaps have been applied.

Training examples contain 2--10 swaps. We evaluate on held-out examples from the same range (ID), immediately beyond the training range with 11--19 swaps, and on longer sequences with 20--40 and 40--80 swaps (OOD).

We compare two input representations. In the \textbf{Template} condition, initial assignments and swap events are serialized using fixed multi-token templates. In the \textbf{Atomic} condition, each initial assignment and swap is represented by a single operation token, such as $[\mathrm{INIT}_{i,c}]$ or $[\mathrm{SWAP}_{a,b}]$. Both representations append five output-slot tokens, one for each entity.
We gather the five slot hidden states after the final LayerNorm and apply a shared
two-layer MLP classifier, $d \rightarrow 2d \rightarrow 5$, independently to each slot.
The classifier parameters are shared across slots.
Detailed tokenization and dataset construction are provided in Appendix A.

\subsection{Model Architectures}
We compare fixed-depth and parameter-shared recurrent Transformers. The \textbf{fixed-depth Transformer} (denoted \textbf{Basic} in our experimental configurations) applies $L$ independently parameterized Transformer blocks, resulting in a fixed amount of sequential computation for every input.

The \textbf{Recurrent Transformer} repeatedly applies a shared Transformer
operator $F_\theta$, following the parameter-sharing paradigm of recurrent-depth
Transformers \citep{dehghani2019universal,fan2025looped}.

\[
H_0 = 0, \qquad H_k = F_\theta\!\left(H_{k-1}+E(x)\right), \qquad k=1,\ldots,K.
\]

Because $F_\theta$ is shared across iterations, the computation budget can grow without introducing new parameters. During training, we set $K=N$, where $N$ is the number of swaps, so recurrent computation increases with problem length. This schedule specifies the computation budget and does not assume that one recurrent iteration corresponds to one individual swap.

We additionally vary positional structure. \textbf{NoPE} uses no explicit positional encoding; \textbf{absolute PE} uses standard sinusoidal positional encoding
\citep{vaswani2017attention}; and \textbf{periodic PE} reuses sinusoidal positional patterns by wrapping the logical positions of swap operations with a fixed period. Full architectural configurations and the precise periodic-PE formulation are given in Appendix A--B.

\subsection{Training and Evaluation}
All models are trained from scratch for 100,000 optimizer steps using seed-deterministic online data generation. Training follows a length curriculum in which the maximum number of swaps gradually increases from 2 to 10, after which examples are sampled across the full 2--10 training range.

Models receive \textbf{final-state supervision only}: cross-entropy loss is applied to the five final entity-state predictions, with no intermediate-state supervision or teacher forcing. The main multiseed comparisons use seeds ${0,1,2}$.

We report \textbf{exact-match accuracy}, which requires all five entity states to be predicted correctly, and \textbf{slot accuracy}, which measures individual entity-state predictions. All accuracy values are reported as percentages; tabular entries are exact match / slot accuracy unless otherwise specified. Unless otherwise specified, recurrent models are evaluated using the same length-dependent computation rule as training, $K=N$. Detailed training, evaluation, and diagnostic protocols are provided in Appendices A and D.

\section{Results and Analysis}

\subsection{Recurrence improves length extrapolation}

\begin{figure}[t]
    \centering
    \includegraphics[width=\linewidth]{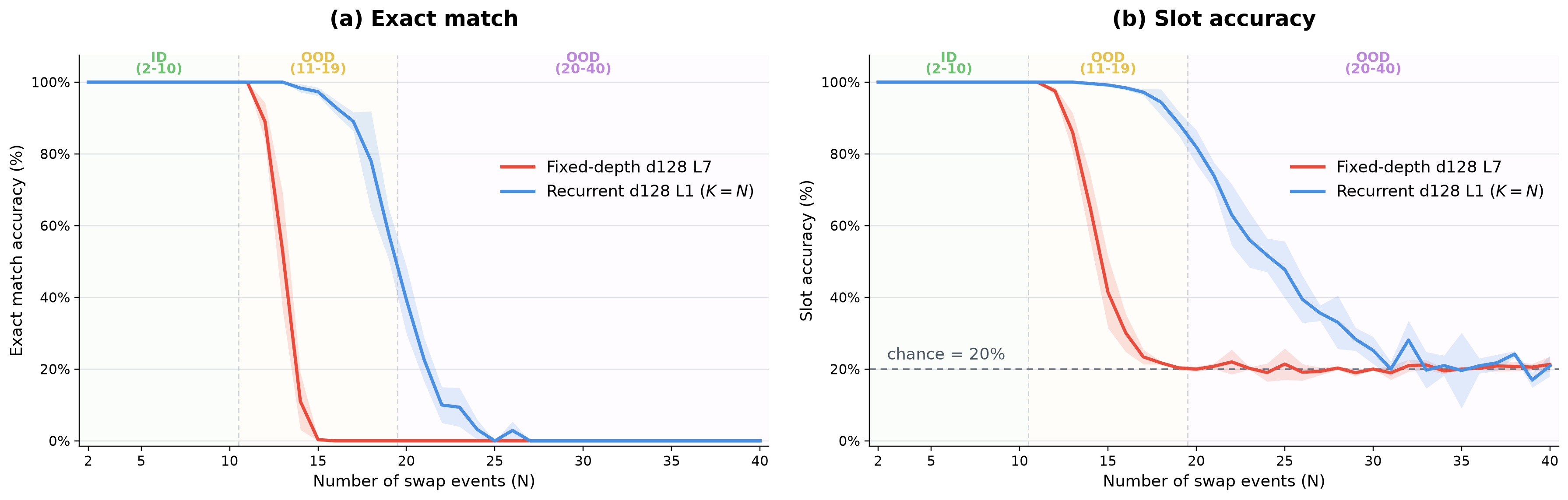}
    \caption{
    Length generalization under Atomic + NoPE. Recurrent d128 L1 with $K=N$ is
    compared with the compute-matched fixed-depth d128 L7 Transformer. Lines show
    means over three seeds, and shaded regions denote $\pm 1$ standard deviation.
    }
    \label{fig:length-generalization}
\end{figure}

We first compare a recurrent Transformer with a fixed-depth Transformer under the same Atomic + NoPE input condition. The recurrent model contains a single shared Transformer block (d128 L1) and is evaluated using the primary length-matched recurrence rule, $K=N$. We compare it against a seven-layer fixed-depth Transformer (Basic d128 L7), which provides the closest expected-forward-compute match over the training-length distribution.

Both models achieve $100\%$ exact-match accuracy on the held-out ID split, confirming that they learn the training task. Their behavior diverges rapidly beyond the training horizon. On OOD 11--19, the recurrent model retains $90.6\%$ exact match and $97.5\%$ slot accuracy, whereas the fixed-depth model reaches $28.1\%$ and $53.9\%$, respectively. On OOD 20--40, the recurrent model still preserves $4.6\%$ exact match and $36.1\%$ slot accuracy, while the fixed-depth model obtains $0.0\%$ exact match and $20.2\%$ slot accuracy. On OOD 40--80, both models exhibit near-zero exact-match accuracy and slot accuracy close to the $20\%$ chance level.

These results show that fitting the ID distribution is not sufficient for length extrapolation. Reusing a shared recurrent operator shifts the generalization boundary substantially beyond that of a fixed-depth model with comparable training-distribution forward compute. However, recurrence does not provide unlimited extrapolation: its advantage decreases with length and vanishes on 40–80 swaps.

Appendix C reports additional recurrent width/depth scaling, deeper fixed-depth baselines, and parameter/compute comparisons. Seed-wise recurrent positional-encoding results are provided in Appendix B.4.

\subsection{Input representation and positional encoding jointly determine extrapolation}
We next perform a 2×2 comparison over input representation (Template or Atomic) and positional encoding (NoPE or absolute sinusoidal PE). This comparison uses the shallow d128 L1 recurrent architecture and the primary K=N protocol. The construction and token-level details of the two input representations are provided in Appendix A.2.

\begin{table}[t]
    \centering
    \resizebox{\linewidth}{!}{%
        \begin{tabular}{llllll}
            \toprule
            Input representation & Positional encoding & ID 2--10 & OOD 11--19 & OOD 20--40 & OOD 40--80 \\
            \midrule
            Template & NoPE
            & 0.0 / 20.1
            & 0.0 / 19.9
            & 0.0 / 19.9
            & 0.0 / 19.0 \\
            
            Template & Absolute sinusoidal
            & 100.0 / 100.0
            & 0.3 / 27.8
            & 0.0 / 20.6
            & 0.0 / 19.8 \\
            
            Atomic & NoPE
            & 100.0 / 100.0
            & \textbf{90.6 / 97.5}
            & \textbf{4.6 / 36.1}
            & 0.1 / 19.1 \\
            
            Atomic & Absolute sinusoidal
            & 100.0 / 100.0
            & 18.0 / 46.6
            & 0.0 / 21.0
            & 0.0 / 20.4 \\
            \bottomrule
        \end{tabular}%
    }
    \caption{Representation $\times$ positional-encoding comparison for the d128 L1 recurrent model under $K=N$. Values are means over seeds 0, 1, and 2, reported as exact match / slot accuracy.}
    \label{tab:representation-pe-comparison}
\end{table}

The two factors jointly shape extrapolation. Template + NoPE fails to learn the ID task, while adding sinusoidal PE solves ID but does not extrapolate beyond the training range. Atomic + NoPE produces the strongest OOD 11--19 and OOD 20–40 results; adding absolute sinusoidal PE to the same Atomic representation substantially reduces this advantage. Thus, recurrent computation is most reusable in this comparison when
operations are represented atomically without being tied to absolute training positions.

Appendix B reports periodic positional-encoding variants and their interaction with recurrent operator depth. Appendix D separately analyzes length-proportional test-time recurrence, where the recurrent budget is scaled beyond the primary $K=N$ rule; this diagnostic post-training analysis is not used to select the primary results.

Taken together, the two main experiments support a conditional view of recurrent length generalization: shared recurrence provides a useful extrapolative bias, while its effectiveness depends strongly on whether operations and positional information are represented in a form that remains reusable beyond the training range.

\subsection{Scaling Recurrent Capacity Further Extends the Extrapolation Horizon}
We finally examine whether increasing model capacity has the same effect for recurrent and fixed-depth architectures. Increasing the capacity of the shared recurrent operator substantially improves extrapolation, whereas scaling the fixed-depth Transformer yields much weaker gains outside the training range.

\begin{table}[H]
    \centering
    \setlength{\abovecaptionskip}{1pt}
    \setlength{\belowcaptionskip}{0pt}
    \begin{tabular}{lll}
        \toprule
        Model & OOD 11--19 & OOD 20--40 \\
        \midrule
        Recurrent d128 L1 & 90.6 / 97.5 & 4.6 / 36.1 \\
        Recurrent d256 L4 & 99.4 / 99.9 & 35.1 / 66.1 \\
        Basic d256 L12 & 38.1 / 63.9 & 0.0 / 20.7 \\
        \bottomrule
    \end{tabular}
    \caption{Recurrent and fixed-depth scaling (exact match / slot accuracy).}
    \label{tab:capacity-scaling}
\end{table}

The recurrent d256 L4 model increases OOD 20--40 exact-match accuracy from 4.6\% to 35.1\%, while the substantially larger fixed-depth d256 L12 model remains at zero exact match on the same split. Thus, simply increasing fixed-depth capacity does not reproduce the extrapolation gains obtained by increasing the capacity of computation that is repeatedly reused.

Even the strongest recurrent models remain near chance on OOD 40--80, indicating that recurrent scaling extends rather than eliminates the extrapolation horizon. Full width/depth scaling results, parameter counts, and compute comparisons are provided in Appendix C.

\section{Conclusion}

Our results show that recurrence can substantially improve length generalization in state tracking, but only under suitable operation representations and positional structure. Scaling recurrent capacity further extends this extrapolation horizon, whereas larger fixed-depth Transformers fail to achieve comparable OOD gains.

Moderate additional test-time recurrence under a length-proportional computation rule can further improve extrapolation, consistent with iterative refinement rather than a strict one-loop-per-operation process. However, the benefit remains finite: performance approaches chance on much longer sequences, and our strongest results rely on atomic representations, leaving the limits of reusable recurrent computation an open question.

\bibliographystyle{plainnat}
\bibliography{references}

\appendix

\section{Experimental Details}

\subsection{State-Tracking Task}

We study a synthetic state-tracking task consisting of five entities and five distinct states. Each example begins with a one-to-one assignment of states to entities, followed by a sequence of pairwise swap operations. Each operation exchanges the states of two entities, and the model predicts the complete assignment after all swaps have been applied. Models receive supervision only on the final state; no intermediate-state labels are used in the main experiments.

Training examples contain 2–10 swaps. We evaluate in-distribution generalization on held-out examples from the same length range, near-range extrapolation on 11–19 swaps, and increasingly longer extrapolation on 20–40 and 40–80 swaps. This controlled construction isolates repeated state-transition composition while providing direct control over the number of state-changing operations.

\begin{center}
\begin{tabular}{lll}
\toprule
Split & Number of swaps & Purpose \\
\midrule
Train & 2–10 & Training distribution \\
ID & 2–10 & Held-out in-distribution evaluation \\
OOD 11--19 & 11–19 & Near-range extrapolation \\
OOD 20–40 & 20–40 & Moderate length extrapolation \\
OOD 40–80 & 40–80 & Long-range extrapolation \\
\bottomrule
\end{tabular}
\end{center}

\subsection{Input Representations}

We compare two representations of state-changing operations. In the \textbf{Template} condition, each assignment and swap is serialized as a multi-token template containing the participating entities, states, and relation tokens. In the \textbf{Atomic} condition, each initial assignment and pairwise swap is represented by a single operation token. Output query slots are represented explicitly in both settings.

This comparison isolates the effect of representing each state transition as a reusable discrete operation rather than as a multi-token template. In particular, the Atomic representation reduces the parsing and binding requirements involved in extracting a state-changing operation from a longer serialization.

\begin{center}
\small
\renewcommand{\arraystretch}{1.15}
\setlength{\tabcolsep}{6pt}

\begin{tabular}{l >{\raggedright\arraybackslash}p{0.76\linewidth}}
\toprule
Representation & Token-level example \\
\midrule

Template &
\texttt{\detokenize{[Entity_0] [has] [the] [blue] [ball] [.] ... 
[Entity_0] [and] [Entity_3] [swapped] [balls] [.] ... 
[SLOT_0] ... [SLOT_4]}} \\[3pt]

Atomic &
\texttt{\detokenize{[INIT_0_BLUE] ... [INIT_4_ORANGE] 
[SWAP_0_3] ... [SLOT_0] ... [SLOT_4]}} \\

\bottomrule
\end{tabular}
\end{center}

The Template row provides an English rendering of the tokenization pattern used
by the generator (initial assignment, swap event, and output slots); the Atomic
row shows the corresponding one-token operation representation. Surface tokens are mapped to vocabulary items before being passed to the model.

\subsection{Model Architectures}

We compare fixed-depth and parameter-shared recurrent Transformers. The \textbf{fixed-depth Transformer} (denoted \textbf{Basic} in our experimental configurations) applies a sequence of independently parameterized Transformer blocks, so its computational depth is fixed once the architecture is specified. The \textbf{Recurrent Transformer} repeatedly applies a shared Transformer operator $F_\theta$. In the main experiments, the input embedding $E$ is reinjected at every recurrent iteration:

\[
H_0 = 0, \qquad H_k = F_\theta\left(H_{k-1}+E\right), \qquad k=1,\ldots,K.
\]

Both architectures use the same causal Transformer-block design within matched width settings; they differ primarily in whether block parameters are independently instantiated across depth or shared across recurrent iterations. The recurrent operator contains $L$ Transformer blocks, all of which are reused at every iteration. During training, recurrent models use $K=N$, where $N$ is the number of swaps in the example.

\begin{center}
\resizebox{\linewidth}{!}{%
\begin{tabular}{ll}
\toprule
Component & Setting \\
\midrule
Self-attention & Causal multi-head self-attention \\
Normalization & LayerNorm \\
Norm placement & Pre-norm \\
Feed-forward network & Two linear layers with GELU activation \\
FFN dimension & $4d$: 512 for $d=128$, 1024 for $d=256$, and 2048 for $d=512$ \\
Dropout & 0.0 \\
Output head & Shared two-layer MLP classifier applied independently to the five output-slot hidden states \\
Input reinjection & $H_{k-1}+E$ at every recurrent iteration \\
\bottomrule
\end{tabular}
}
\end{center}

\begin{center}
\begin{tabular}{llll}
\toprule
Architecture & Parameterization & Executed depth & Train-time computation \\
\midrule
Basic & Independent blocks & Fixed $L$ & Fixed \\
Recurrent & Shared $L$-block operator $F_\theta$ & $L\times K$ & $K=N$ \\
\bottomrule
\end{tabular}
\end{center}

The parameter counts of the recurrent configurations used in the scaling analysis are reported in Appendix C.1.

\subsection{Training Protocol}

All models are trained from scratch on synthetically generated examples. The final main experiments use seed-deterministic online generation and a step-to-maximum-length curriculum. At zero-indexed optimizer step $s$, the maximum allowed number of swaps is

\[
N_{\max}(s)=\min\!\left(10,\,2+\left\lfloor\frac{s}{1000}\right\rfloor\right),
\]

and the training length is sampled as $N\sim\operatorname{UniformInteger}(2,N_{\max}(s))$. A single value of $N$ is sampled using the deterministic random-number generator at each optimizer step and is shared by all examples in that step’s batch. Thus, the maximum length increases by one every 1,000 optimizer steps until reaching 10 at step 8,000; training thereafter samples uniformly from 2–10 swaps through step 99,999. For a recurrent model, the number of training iterations is matched to the sampled number of swaps, $K=N$.

Training supervision is applied only to the final state using cross-entropy loss. No intermediate-state supervision or teacher forcing is used in the main experiments.

\begin{center}
\begin{tabular}{ll}
\toprule
Setting & Value \\
\midrule
Training data & Seed-deterministic online generation \\
Training length & 2–10 swaps \\
Curriculum & Step-to-maximum-length; $N\sim\operatorname{UniformInteger}(2,N_{\max}(s))$ \\
Optimizer steps & 100k \\
Recurrent iterations & $K=N$ \\
Supervision & Final-state cross-entropy only \\
Attention & Causal \\
Optimizer & AdamW \\
Learning rate & $3\times10^{-4}$ \\
Batch size & 128 \\
Weight decay & 0.01 \\
Dropout & 0.0 \\
\bottomrule
\end{tabular}
\end{center}

\subsection{Compute Resources}

Experiments were run as single-GPU jobs on NVIDIA RTX 3090 GPUs with 24 GB memory. Based on recorded wall-clock training times where available, a typical Recurrent d128 L1, Recurrent d256 L4, and Basic d256 L12 training run required approximately 0.8, 2.0, and 1.0 GPU-hours, respectively. The d256 L4 value is an approximate estimate obtained by scaling the observed d256 L2 recurrent timing by executed depth, since raw timing logs for that condition were not retained in this repository. The final reported experiments used approximately 80 GPU-hours in total, including estimated timing for reported runs without retained logs. Additional exploratory experiments required approximately 10 GPU-hours.

\subsection{Reproducibility Details}

\paragraph{Implementation details.}
All Transformer blocks use multi-head scaled dot-product self-attention. The number of attention heads is the $h$ value reported in each $d/h/\mathrm{ff}$ configuration: $d=128$ models use 4 heads, $d=256$ models use 8 heads, and $d=512$ models use 8 heads. This gives head dimensions of 32, 32, and 64, respectively. Each attention layer uses a learned $D \rightarrow 3D$ QKV projection and a learned $D \rightarrow D$ output projection. All main models use causal attention.

\paragraph{Output classifier.}
After the final LayerNorm, we gather the five output-slot hidden states and apply the same slot classifier independently to each slot. The classifier is a shared two-layer MLP,
\[
g(h)=W_2\,\mathrm{ReLU}(W_1 h+b_1)+b_2.
\]
With model width $D$, the classifier has structure $D \rightarrow 2D \rightarrow 5$, with biases enabled in both linear layers. Thus, the $d=128$ setting uses $\mathrm{Linear}(128,256)$, ReLU, and $\mathrm{Linear}(256,5)$. The classifier parameters are shared across the five output slots within each model.

\paragraph{Evaluation data.}
Evaluation uses fixed synthetic JSONL splits generated deterministically and reused across model seeds. The ID split contains 500 examples with 2--10 swaps. The OOD 11--19 (Boundary) split contains 900 examples with 11--19 swaps, exactly 100 examples per swap length. The OOD 20--40 and OOD 40--80 splits each contain 500 examples. The optional Easy $N=2$ diagnostic split contains 500 examples.

\paragraph{Random seeds and data generation.}
The main results of multiseed use model/training seeds 0, 1, and 2. The implementation seeds Python \texttt{random}, PyTorch, and CUDA random-number generators. Online training examples are generated deterministically from the tuple $(\mathrm{seed}, \mathrm{optimizer\_step}, \mathrm{sample\_index})$ using a SplitMix-style 64-bit seed mixer. Evaluation splits are generated by sampling random initial color permutations and random swap pairs, replaying the symbolic swaps to compute labels, and deduplicating examples by initial assignment and swap sequence. The evaluation split seeds are 1000 for ID, 0 for OOD 11--19, 2004 for OOD 20--40, 2008 for OOD 40--80, and 123 for the Easy $N=2$ diagnostic split.

\paragraph{Software.}
The repository specifies Python $\geq 3.10$. The locked environment resolves PyTorch to version 2.13.0, and experiments use CUDA 13.0. Other locked major dependencies include PyYAML 6.0.3, tqdm 4.70.0, and pytest 9.1.1. NumPy is resolved according to the Python-version markers in the lockfile.

\paragraph{Statistical reporting.}
Unless otherwise specified, multiseed results are reported over three
independent training seeds (0, 1, and 2). Variability is reported as
$\pm 1$ sample standard deviation across seeds, computed with denominator
$n-1$ (\texttt{ddof=1}).

\section{Extended Positional-Encoding Results}

\subsection{Positional-Encoding Variants}

Under the Atomic representation, we compare \textbf{NoPE}, standard \textbf{absolute sinusoidal positional encoding}, and \textbf{Periodic positional encoding}. Periodic PE does not introduce a new sinusoidal function. Instead, it wraps the logical position indices assigned to swap tokens with period $P$, after which the standard sinusoidal encoding is applied to the wrapped indices. This allows the same positional patterns to recur beyond the training range.

For the $t$-th swap token, its wrapped logical position is

\[
\tilde p_t = p_{\mathrm{swap}} + ((t-1) \bmod P),
\]

where $p_{\mathrm{swap}}$ denotes the first logical swap position. Non-swap tokens retain their designated logical positions. The standard sinusoidal encoding is then computed from $\tilde p$:

\[
\operatorname{PE}(\tilde p,2i)
= \sin\left(\tilde p / 10000^{2i/d}\right),
\qquad
\operatorname{PE}(\tilde p,2i+1)
= \cos\left(\tilde p / 10000^{2i/d}\right).
\]

We evaluate $P\in\{1,2\}$ at shallow recurrent depth (L1) and $P\in\{1,2,5\}$ at L6. All aggregate $K=N$ results reported for these conditions are means over seeds 0, 1, and 2.

\subsection{Positional-Encoding Results at Shallow Recurrent Depth}

Table B.1 reports seed 0/1/2 means for the d128 L1 recurrent model under the primary $K=N$ protocol. Additional test-time recurrence analyses are reported separately in Appendix D.

\textbf{Table B.1: Atomic d128 L1 positional-encoding comparison.}

\begin{center}
\resizebox{\linewidth}{!}{%
\begin{tabular}{lllll}
\toprule
Positional encoding & ID $K=N$ & OOD 11--19 $K=N$ & OOD 20–40 $K=N$ & OOD 40–80 $K=N$ \\
\midrule
NoPE
& 100.0 / 100.0
& \textbf{90.6 / 97.5}
& \textbf{4.6 / 36.1}
& 0.1 / 19.1 \\

Absolute sinusoidal
& 100.0 / 100.0
& 18.0 / 46.6
& 0.0 / 21.0
& 0.0 / 20.4 \\

Periodic $P=1$
& 100.0 / 100.0
& 81.7 / 92.7
& 4.5 / 32.9
& 0.0 / 20.1 \\

Periodic $P=2$
& 100.0 / 100.0
& 87.1 / 96.4
& 3.9 / 36.0
& 0.1 / 19.9 \\
\bottomrule
\end{tabular}
}
\end{center}

At shallow recurrent depth, NoPE gives the strongest extrapolation. Absolute sinusoidal encoding sharply degrades performance, whereas periodic encoding recovers much of the NoPE result. Thus, the failure of absolute PE does not imply that all positional structure is incompatible with length extrapolation.

\subsection{Interaction with Recurrent Depth}

The effect of positional structure changes with recurrent operator depth. At L1, NoPE provides the strongest extrapolation. At L6, however, Periodic $P=2$ produces the strongest mean OOD 11--19 and OOD 20--40 performance among the compared positional conditions. Table B.2 reports seed 0/1/2 means under the primary $K=N$ protocol.

\textbf{Table B.2: Atomic d128 L6 positional-encoding comparison under $K=N$.}

\begin{center}
\resizebox{\linewidth}{!}{%
\begin{tabular}{llll}
\toprule
Positional encoding & OOD 11--19 $K=N$ & OOD 20--40 $K=N$ & OOD 40--80 $K=N$ \\
\midrule
NoPE
& 97.5 / 99.3
& 19.3 / 52.5
& 0.0 / 20.3 \\

Periodic $P=1$
& 98.0 / 99.4
& 19.9 / 54.0
& 0.1 / 20.4 \\

Periodic $P=2$
& \textbf{99.0 / 99.7}
& \textbf{27.3 / 62.0}
& 0.1 / 21.6 \\

Periodic $P=5$
& 49.5 / 77.2
& 2.9 / 31.0
& 0.0 / 20.2 \\
\bottomrule
\end{tabular}
}
\end{center}

These results suggest an interaction between positional inductive bias and recurrent capacity rather than a universally optimal positional scheme. Periodic structure is not uniformly beneficial: $P=2$ performs best at L6, whereas $P=5$ substantially degrades extrapolation. Because increasing operator depth also changes the computation performed at each recurrent iteration, the mechanism underlying this interaction remains unresolved.

\subsection{Seed-Wise Positional-Encoding Results}

Table B.3 reports the individual seed results for the three primary L6 positional-encoding conditions used in the seed-variability analysis.

\textbf{Table B.3: Seed-wise d128 L6 recurrent results under $K=N$.}

\begin{center}
\begin{tabular}{lllll}
\toprule
Condition & Seed & OOD 11--19 & OOD 20--40 & OOD 40--80 \\
\midrule
NoPE & 0 & 99.0 / 99.8 & 29.4 / 62.0 & 0.0 / 20.8 \\
     & 1 & 97.7 / 99.4 & 17.8 / 53.4 & 0.0 / 19.5 \\
     & 2 & 95.8 / 98.8 & 10.8 / 42.1 & 0.0 / 20.6 \\

Periodic $P=1$ & 0 & 98.6 / 99.6 & 20.6 / 56.1 & 0.0 / 21.8 \\
               & 1 & 97.3 / 99.2 & 17.8 / 50.8 & 0.0 / 19.4 \\
               & 2 & 98.2 / 99.5 & 21.2 / 55.2 & 0.2 / 20.0 \\

Periodic $P=2$ & 0 & 98.6 / 99.6 & 23.6 / 59.7 & 0.2 / 22.3 \\
               & 1 & 99.8 / 100.0 & 35.2 / 67.8 & 0.2 / 21.3 \\
               & 2 & 98.7 / 99.6 & 23.0 / 58.4 & 0.0 / 21.3 \\
\bottomrule
\end{tabular}
\end{center}

On OOD 20--40, the corresponding mean $\pm$ sample standard deviation for exact match / slot accuracy is $19.3\pm9.4$ / $52.5\pm10.0$ for NoPE, $19.9\pm1.8$ / $54.0\pm2.8$ for Periodic $P=1$, and $27.3\pm6.9$ / $62.0\pm5.1$ for Periodic $P=2$.

\section{Scaling and Baseline Details}

\subsection{Recurrent Scaling at \texorpdfstring{$K=N$}{K=N}}

We vary the width and depth of the shared recurrent operator to test whether increasing recurrent capacity extends length extrapolation. All values in Table C.1 are seed 0/1/2 means under the primary $K=N$ protocol.

\textbf{Table C.1: Recurrent NoPE scaling.}

\begin{center}
\resizebox{\linewidth}{!}{%
\begin{tabular}{lllll}
\toprule
Model & Parameters & OOD 11--19 $K=N$ & OOD 20–40 $K=N$ & OOD 40–80 $K=N$ \\
\midrule
d128 L1 & 240,005
& 90.6 / 97.5 & 4.6 / 36.1 & 0.1 / 19.1 \\

d128 L2 & 438,277
& 96.1 / 99.0 & 12.3 / 46.4 & 0.0 / 20.8 \\

d128 L4 & 834,821
& 98.2 / 99.5 & 23.1 / 56.7 & 0.1 / 21.2 \\

d128 L6 & 1,231,365
& 97.5 / 99.3 & 19.3 / 52.5 & 0.0 / 20.3 \\

d256 L1 & 938,757
& 96.4 / 99.0 & 12.1 / 45.0 & 0.1 / 20.9 \\

d256 L2 & 1,728,517
& 98.7 / 99.7 & 24.3 / 58.2 & 0.0 / 20.6 \\

d256 L4 & 3,308,037
& \textbf{99.4 / 99.9}
& \textbf{35.1 / 66.1}
& 0.2 / 22.0 \\

d512 L1 & 3,712,517
& 98.0 / 99.5 & 15.1 / 47.7 & 0.0 / 20.4 \\
\bottomrule
\end{tabular}
}
\end{center}

Increasing recurrent capacity substantially improves OOD 11--19 and OOD 20–40 performance. The gains are not monotonic in parameter count alone: widening a shallow operator does not fully substitute for increasing recurrent operator depth. Even the strongest models, however, remain near chance on OOD 40–80.

\subsection{Fixed-Depth Scaling}

To distinguish recurrent computation from generic model scaling, we scale fixed-depth Basic Transformers under the same state-tracking formulation. Increasing fixed depth solves the in-distribution task and improves OOD 11--19 accuracy, but it does not yield comparable extrapolation on OOD 20--40. All rows in Table C.2 report means over seeds 0, 1, and 2.

\textbf{Table C.2: Fixed-depth Basic Transformer results.}

\begin{center}
\resizebox{\linewidth}{!}{%
\begin{tabular}{llllll}
\toprule
Model & Seeds & ID & OOD 11--19 & OOD 20--40 & OOD 40--80 \\
\midrule
NoPE d128 L1 & 0/1/2 mean
& 9.9 / 42.8
& 0.1 / 20.9
& 0.0 / 20.2
& 0.1 / 20.4 \\

NoPE d128 L6 & 0/1/2 mean
& 100.0 / 100.0
& 24.8 / 49.9
& 0.0 / 20.1
& 0.0 / 20.2 \\

NoPE d128 L7 & 0/1/2 mean
& 100.0 / 100.0
& 28.1 / 53.9
& 0.0 / 20.2
& 0.0 / 20.1 \\

NoPE d128 L10 & 0/1/2 mean
& 100.0 / 100.0
& 33.2 / 58.4
& 0.0 / 20.3
& 0.0 / 19.7 \\

NoPE d256 L12 & 0/1/2 mean
& 100.0 / 100.0
& 38.1 / 63.9
& 0.0 / 20.7
& 0.0 / 19.8 \\

Periodic $P=2$ d128 L1 & 0/1/2 mean
& 15.3 / 45.1
& 0.0 / 20.5
& 0.0 / 19.8
& 0.1 / 20.0 \\

Periodic $P=2$ d128 L6 & 0/1/2 mean
& 100.0 / 100.0
& 18.9 / 48.4
& 0.0 / 19.9
& 0.0 / 19.9 \\

Periodic $P=2$ d128 L10 & 0/1/2 mean
& 100.0 / 100.0
& 33.0 / 60.6
& 0.0 / 20.3
& 0.0 / 20.3 \\
\bottomrule
\end{tabular}
}
\end{center}

The compute-matched NoPE d128 L7 baseline reaches 28.1\% OOD 11--19 exact-match accuracy but remains at 0.0\% exact match on OOD 20--40. A substantially larger Basic d256 L12 model improves OOD 11--19 exact match to 38.1\%, yet likewise remains at 0.0\% exact match on OOD 20--40. In contrast, the recurrent d256 L4 model reaches 35.1\% exact match on the same OOD split.

\subsection{Parameter and Compute Comparison}

Recurrent models reuse the parameters of the same operator across iterations, whereas Basic Transformers instantiate independently parameterized blocks. We therefore distinguish parameter count from executed computation.

For the train distribution $N\in[2,10]$, an expected-forward-FLOPs analysis identifies Basic d128 L7 as the closest fixed-depth compute comparison for Recurrent d128 L1.

\textbf{Table C.3: Expected train-distribution forward compute.}

\begin{center}
\begin{tabular}{lll}
\toprule
Model & Expected forward FLOPs per sample & Relative to recurrent \\
\midrule
Recurrent d128 L1, $K=N$ & 41.76M & 1.000$\times$ \\
Basic d128 L6 & 38.46M & 0.921$\times$ \\
Basic d128 L7 & 44.82M & 1.073$\times$ \\
\bottomrule
\end{tabular}
\end{center}

The fixed-depth scaling and compute-matched comparisons jointly show that increasing independently parameterized depth does not reproduce the OOD gains obtained by repeatedly applying shared computation.

\subsection{Variability of Main Scaling Comparisons}

Table C.4 reports sample standard deviations across seeds 0, 1, and 2
for the principal scaling comparison. Values are reported as mean $\pm$ sample standard deviation in percentage points for exact match / slot accuracy.

\textbf{Table C.4: Variability of the principal scaling results.}

\begin{center}
\begin{tabular}{lll}
\toprule
Model & OOD 11--19 & OOD 20--40 \\
\midrule
Recurrent d128 L1
& $90.6 \pm 2.7$ / $97.5 \pm 0.9$
& $4.6 \pm 1.1$ / $36.1 \pm 2.7$ \\

Recurrent d256 L4
& $99.4 \pm 0.3$ / $99.9 \pm 0.1$
& $35.1 \pm 0.5$ / $66.1 \pm 1.0$ \\

Basic d256 L12
& $38.1 \pm 4.5$ / $63.9 \pm 4.6$
& $0.0 \pm 0.0$ / $20.7 \pm 0.4$ \\
\bottomrule
\end{tabular}%
\end{center}

\section{Diagnostic Test-Time Recurrence Analysis}
Across multiple recurrent configurations, predefined fixed-$K$ sweeps show
that additional test-time recurrence beyond the primary $K=N$ rule often
improves Boundary and OOD 20--40 performance. This pattern appears across
multiple model sizes and positional-encoding conditions, although the optimal
recurrence budget varies by configuration and the gains eventually saturate.

We additionally examine this effect under a length-proportional computation
rule using a Recurrent d256 L2 model with the Atomic+NoPE configuration
(seed 0), trained for 100,000 optimizer steps under the same primary $K=N$
training rule.

Rather than assigning the same fixed recurrence budget to examples of different
lengths, we scale the number of recurrent iterations proportionally to the
number of swaps. For an example containing $N$ swaps, we set
\[
K=\lceil \alpha N\rceil,
\]
where $\alpha$ controls the amount of test-time computation relative to the
training-time rule $K=N$.

\textbf{Table D.1: Length-proportional test-time recurrence for the Recurrent d256 L2
Atomic+NoPE model (seed 0).}

\begin{center}
\begin{tabular}{llll}
\toprule
Recurrence multiplier $\alpha$ & OOD 11--19 & OOD 20--40 & OOD 40--80 \\
\midrule
1.00 & 98.6 / 99.6 & 20.0 / 53.6 & 0.0 / 20.0 \\
1.25 & \textbf{98.9 / 99.7} & \textbf{29.4} / 61.1 & 0.0 / \textbf{22.3} \\
1.50 & 98.8 / 99.7 & 29.2 / \textbf{61.2} & \textbf{0.2} / 21.6 \\
2.00 & 98.4 / 99.6 & 28.0 / 60.0 & 0.2 / 20.9 \\
3.00 & 97.4 / 99.4 & 24.4 / 58.7 & 0.0 / 20.0 \\
\bottomrule
\end{tabular}
\end{center}

Values are reported as exact match / slot accuracy. Moderate additional recurrence improves extrapolation beyond the primary $K=N$ rule. On OOD 20--40, exact-match accuracy increases from 20.0\% at $\alpha=1$ to 29.4\% at $\alpha=1.25$, while slot accuracy reaches 61.2\% at $\alpha=1.5$. Performance subsequently declines as the recurrence multiplier increases further, indicating that additional computation is beneficial only within a limited range rather than monotonically improving performance.

The same pattern is weaker on OOD 11--19 because performance is already near saturation under $K=N$. In contrast, OOD 40--80 remains near chance for all tested recurrence multipliers. Although slot accuracy increases slightly for $\alpha=1.25$ and $\alpha=1.5$, the effect is small and does not constitute meaningful long-range generalization.

These results support an iterative-refinement interpretation of recurrent computation. The model can benefit from running beyond the length-matched $K=N$ budget, but excessive recurrence eventually reduces performance, and additional test-time computation alone does not eliminate the long-range extrapolation limit.

\end{document}